\documentclass[sigconf]{acmart}
\usepackage{algorithm}
\usepackage{algorithmic}
\usepackage{graphicx}
\usepackage{subcaption}
\usepackage{balance}
\newcommand{\tra}{record}
\newcommand{\tras}{records}
\newcommand{\Tra}{Record}

\newcommand{\name}{FATA-Trans}

\AtBeginDocument{%
  \providecommand\BibTeX{{%
    \normalfont B\kern-0.5em{\scshape i\kern-0.25em b}\kern-0.8em\TeX}}}

\copyrightyear{2023}
\acmYear{2023}
\setcopyright{acmlicensed}\acmConference[CIKM '23]{Proceedings of the 32nd
ACM International Conference on Information and Knowledge
Management}{October 21--25, 2023}{Birmingham, United Kingdom}
\acmBooktitle{Proceedings of the 32nd ACM International Conference on
Information and Knowledge Management (CIKM '23), October 21--25, 2023,
Birmingham, United Kingdom}
\acmPrice{15.00}
\acmDOI{10.1145/3583780.3614879}
\acmISBN{979-8-4007-0124-5/23/10}

\begin{document}

\title{FATA-Trans: \underline{F}ield \underline{A}nd \underline{T}ime-\underline{A}ware \underline{Trans}former for Sequential Tabular Data}


\author{Dongyu Zhang}
\email{dzhang5@wpi.edu}
\affiliation{
\institution{Worcester Polytechnic Institute}
\city{Worcester}
  \state{Massachusetts}
  \country{USA}
}

\author{Liang Wang}
\email{liawang@visa.com}
\affiliation{
\institution{Visa Research}
\city{Palo Alto}
  \state{California}
  \country{USA}
}

\author{Xin Dai}
\email{xidai@visa.com}
\affiliation{
\institution{Visa Research}
\city{Palo Alto}
  \state{California}
  \country{USA}
}

\author{Shubham Jain}
\email{shubhjai@visa.com}
\affiliation{
\institution{Visa Research}
\city{Palo Alto}
  \state{California}
  \country{USA}
}

\author{Junpeng Wang}
\email{junpenwa@visa.com}
\affiliation{
\institution{Visa Research}
\city{Palo Alto}
  \state{California}
  \country{USA}
}

\author{Yujie Fan}
\email{yufan@visa.com}
\affiliation{
\institution{Visa Research}
\city{Palo Alto}
  \state{California}
  \country{USA}
}

\author{Chin-Chia Michael Yeh}
\email{miyeh@visa.com}
\affiliation{
\institution{Visa Research}
\city{Palo Alto}
  \state{California}
  \country{USA}
}

\author{Yan Zheng}
\email{yazheng@visa.com}
\affiliation{
\institution{Visa Research}
\city{Palo Alto}
  \state{California}
  \country{USA}
}

\author{Zhongfang Zhuang}
\email{zzhuang@visa.com}
\affiliation{
\institution{Visa Research}
\city{Palo Alto}
  \state{California}
  \country{USA}
}

\author{Wei Zhang}
\email{wzhan@visa.com}
\affiliation{
\institution{Visa Research}
\city{Palo Alto}
  \state{California}
  \country{USA}
}

\renewcommand{\shortauthors}{Dongyu Zhang, et al.}

\begin{abstract}
 Sequential tabular data is one of the most commonly used data types in real-world applications. Different from conventional tabular data, where rows in a table are independent, sequential tabular data contains rich contextual and sequential information, where some fields are \textit{dynamically} changing over time and others are \textit{static}. 
Existing transformer-based approaches analyzing sequential tabular data overlook the differences between dynamic and static fields by replicating and filling static fields into each \tra{}, and ignore temporal information between rows, which leads to three major disadvantages: (1) computational overhead, (2) artificially simplified data for masked language modeling pre-training task that may yield less meaningful representations, and (3) disregarding the temporal behavioral patterns implied by time intervals. 
In this work, we propose FATA-Trans, a model with two field transformers for modeling sequential tabular data, where each processes static and dynamic field information separately.  
\name\ is \textit{field}- and \textit{time}-aware for sequential tabular data. 
The \textit{field}-type embedding in the method enables FATA-Trans to capture differences between static and dynamic fields. 
The \textit{time}-aware position embedding exploits both order and time interval information between rows, which helps the model detect underlying temporal behavior in a sequence.
Our experiments on three benchmark datasets demonstrate that the learned representations from FATA-Trans consistently outperform state-of-the-art solutions in the downstream tasks. 
We also present visualization studies to highlight the insights captured by the learned representations, enhancing our understanding of the underlying data. Our codes are available at \url{https://github.com/zdy93/FATA-Trans}.
\end{abstract}





\maketitle

\section{Introduction}
Sequential tabular data is one of the most commonly used data types in real-world applications such as medical diagnosis\cite{zhang2020time}, recommendation systems\cite{zhang2019deep}, click-through-rate (CTR) prediction\cite{muhamed2021ctr}, and transaction anomaly detection\cite{zhang2021transaction}.
In a sequential tabular dataset, there are multiple rows and columns, where each row corresponds to a \textit{record} and each column represents a \textit{field}.
An example of sequential tabular data is the credit card transaction data which captures purchasing activities of cardholders over time. 
Another example is the user review data from a website, which logs users’ comments and ratings for the products they rented or bought in the past.

\begin{figure}
    \centering
    \includegraphics[width=1\columnwidth]{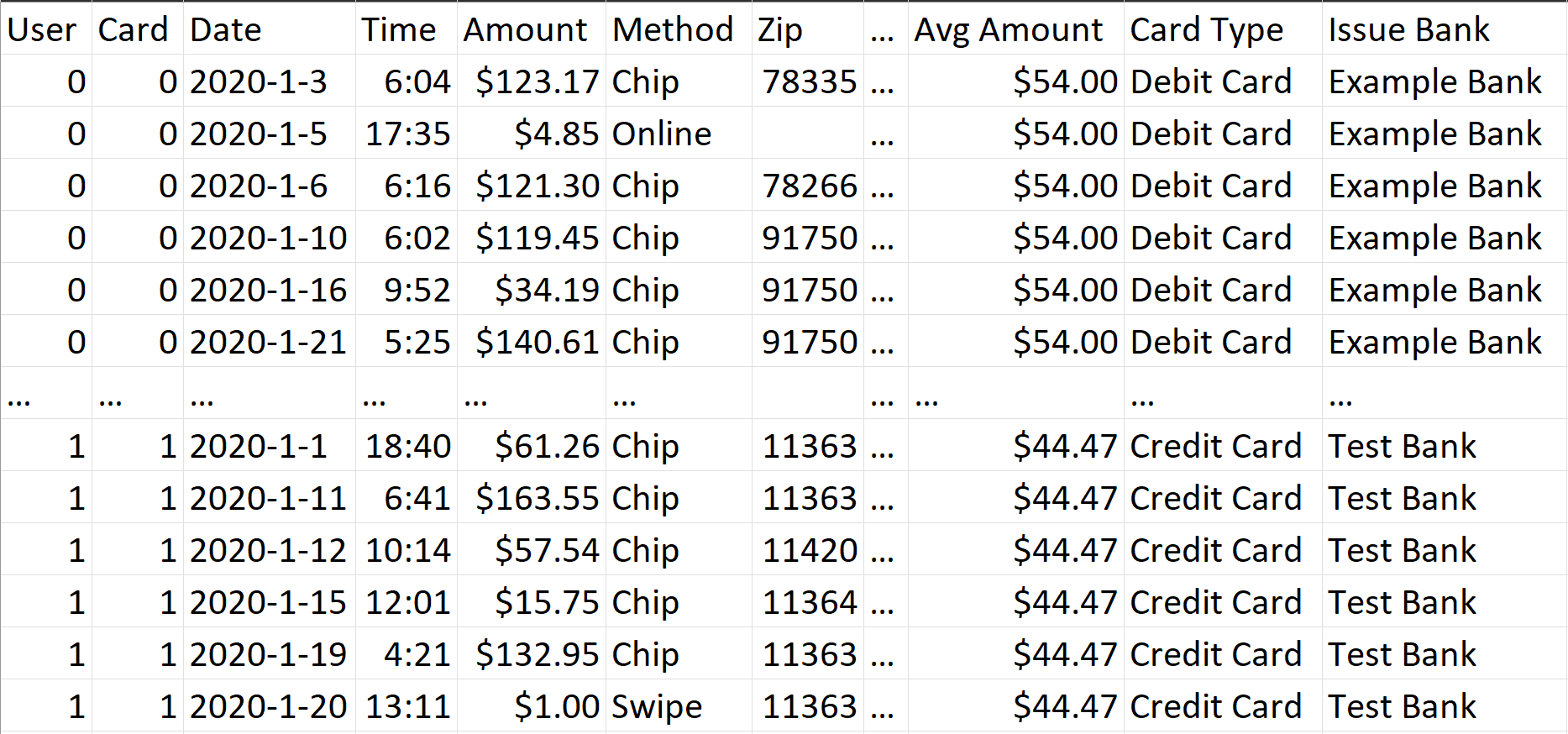}
    \caption{A screenshot of a synthetic sequential tabular dataset showing transaction records. Each row represents a transaction associated with a user and card identifier (columns 1 and 2). A user can own multiple cards. The dataset contains dynamic fields (columns 3 to 7) that vary across transactions and static fields (last three columns) that remain constant. The dataset includes two distinct sequences of transactions: the first six rows and the last six rows.
    }
    \label{fig:tabular_screenshot}
\end{figure}

Figure \ref{fig:tabular_screenshot} depicts a screenshot of a modified sequential tabular dataset derived from a synthetic transaction dataset created by \cite{altman2021synthesizing, padhi2021tabular}. Each row in the dataset represents a transaction record. The first six rows form a transaction sequence associated with User 0's card 0, while the last six rows represent another sequence linked to User 1's card 1. Within these transaction sequences, certain fields in the data are \textit{dynamic}, capturing a user's \textit{local} activities that are transient in nature. Conversely, other fields are \textit{static}, reflecting a user's \textit{global} identity that remains stable over time. For example, in Figure 1, dollar amounts are dynamic as they tend to vary from one transaction to another. On the other hand, card type (e.g., Debit Card) and issuer bank (e.g., Example Bank) remain constant throughout the transactions. Both dynamic and static fields play distinct but significant roles within a transaction sequence.

The concept of dynamic and static fields extends beyond the raw fields present in transaction data and can also include derived fields or features. For example, a derived static field could be the average dollar amount of each user's transactions over a specific time period. In the context of user review data, the ratings given by a user to various products can be considered dynamic since they can change over time. However, the user's average rating over a previous time period would be considered a static field, as it represents an aggregated value that remains constant within that specific period.

Furthermore, in sequential tabular datasets, capturing time and order information is crucial for understanding user behaviors. Figure 2 provides an example of transaction sequences associated with three different cardholders. Although the field values across the three sequences are identical, the timing and order of the transactions differ. The presence of time and order information becomes critial in identifying abnormal transaction behaviors. In the given example, the first sequence exhibits no abnormal transactions, but the other sequences contain abnormal transactions. Hence, considering time and order information becomes essential during model training to effectively capture user behaviors within sequential tabular datasets.

\begin{figure}
\includegraphics[width=1\columnwidth]{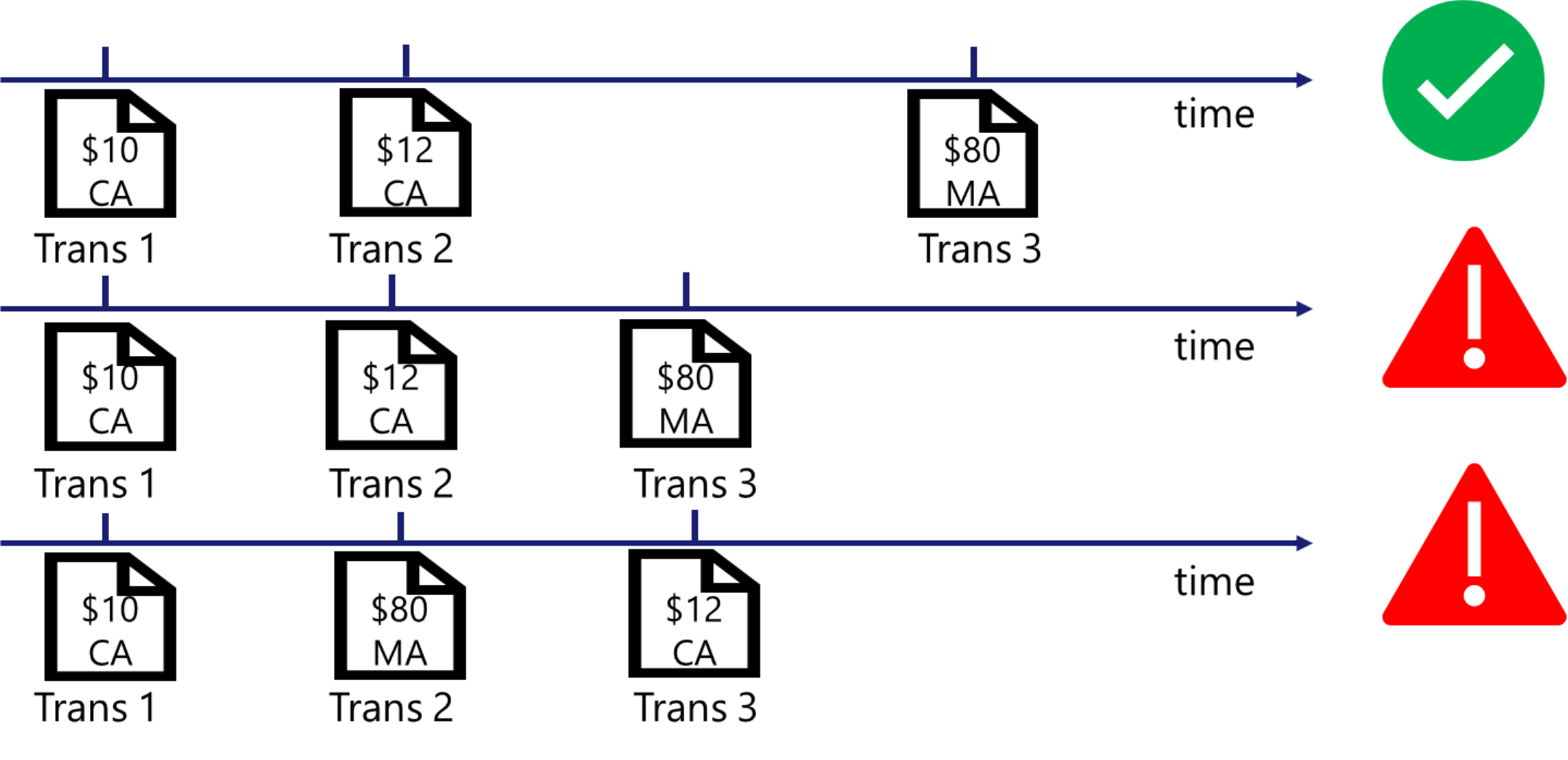}
\caption{
Example of transaction record sequences with the same field values but different order and timing. Abnormal transactions in sequences 2 and 3 cannot be detected based solely on field values. This highlights the significance of time and order information in detecting abnormal transaction patterns.}
\label{fig:time_and_order}
\end{figure}

Extensive research has been dedicated to modeling sequential tabular data. In the early stage, recurrent neural networks (RNNs) were employed and demonstrated outstanding performance\cite{hidasi2015session,zhang2014sequential,yu2016dynamic}. However, more recent efforts have primarily focused on transformer-based models due to their superior ability to capture long-range dependencies and effectively handle sequential data.

The pioneering work of Kang and McAuley\cite{kang2018self} has inspired numerous subsequent studies, leading to the development of various innovative transformer-based model architectures specifically tailored for modeling sequential tabular data \cite{sun2019bert4rec,chen2022time,chen2022denoising,zhang2019feature,chen2019behavior,wu2020sse,zhou2020s3,padhi2021tabular, de2021transformers4rec}. Among these architectures, TabBERT, initially developed by Padhi et al.\cite{padhi2021tabular} and further enhanced in subsequent works\cite{han2022luna,luetto2023one}, presents a powerful and comprehensive framework for processing sequential tabular data. TabBERT is a pre-trained transformer architecture trained using masked language modeling (MLM)\cite{devlin2018bert}, where it predicts masked tokens. It adopts a hierarchical approach to encode a series of transactions using two transformers. At the first level, the transformer processes individual tabular rows (transactions) by considering each field value as a token, thereby generating transaction embeddings. The second-level transformer takes these transaction embeddings as input and produces sequence embeddings.

While TabBERT is applicable to a wide range of sequential tabular data, it has two limitations. First, although it can accommodate both dynamic and static fields, it does not differentiate between them. Static fields are replicated in every record within a sequence, leading to computational overhead when multiple static fields or lengthy sequences are present. Additionally, since TabBERT employs MLM to predict masked tokens, predicting a masked static field in one record becomes relatively easy given the same field in other records. This can potentially hinder the model's ability to learn meaningful representations for sequential tabular data during pre-training. Second, although TabBERT leverages order information through the position embedding layer, it fails to consider the critical time information necessary to capture important user behavior patterns in a sequence, as depicted in Figure 2. It is important to note that these limitations are not exclusive to TabBERT; they also exist in other recently proposed transformer-based architectures dealing with multivariate sequences\cite{zhang2019feature,chen2019behavior,wu2020sse,zhou2020s3}.

In general, previous methods have the following limitations:
\begin{itemize}
    \item \textit{Failure to distinguish static and dynamic fields}: Static and dynamic fields in sequential tabular datasets have distinct roles and impacts. However, previous works did not differentiate between these two types of fields, leading to an inability to capture their unique characteristics.
    \item \textit{Negative effect caused by static field replication}: Previous approaches replicated static fields across records, resulting in unnecessary computational overhead and an excessive reduction in the complexity of the model's pre-training task.
    \item \textit{Failure to utilize both time and order information}: Time and order information are crucial in tasks involving sequential tabular data. However, previous methods either neglected or only partially incorporated the influence of time and order information. A more effective design is necessary to fully leverage the importance of both time and order information. 
\end{itemize}

To address these limitations, we present FATA-Trans, an innovative transformer-based architecture designed to handle sequential tabular datasets while considering field and time information. Our proposed method employs a hierarchical approach for processing input record sequences. At the first level, our method incorporates two field transformers: a static field transformer and a dynamic field transformer. These transformers independently encode the static fields and dynamic fields within each record, resulting in a static field embedding and a series of dynamic field embeddings.  At the second level of the architecture, the generated embeddings from the first level are utilized to create sequence embeddings. Here, a field-type embedding is introduced to discern between static and dynamic fields. Additionally, a time-aware position embedding is employed in the second-level transformer to capture time and order information.

The contributions of our paper are as follows: We propose \name{}, an innovative transformer-based architecture that learns representations of sequential tabular datasets. The architecture consists of three key components: (1) two field transformers at the first level that process static and dynamic fields separately, (2) a field-type embedding in the second-level transformer that can distinguish static and dynamic \tra{} embeddings, and (3) a customized time-aware position embedding that considers the impact of both time and order of \tras{} in a sequence. 

Our experimental results demonstrate that the learned embeddings from FATA-Trans consistently yield improved performance compared to state-of-the-art solutions. Notably, the pre-training process of FATA-Trans outperforms TabBERT in terms of speed. Furthermore, we employed visualization techniques to explore the extracted embeddings from the pre-trained model, uncovering meaningful patterns within the data.

\section{Methodology}
\subsection{Problem Definition}
\begin{figure}
\includegraphics[width=1\columnwidth]{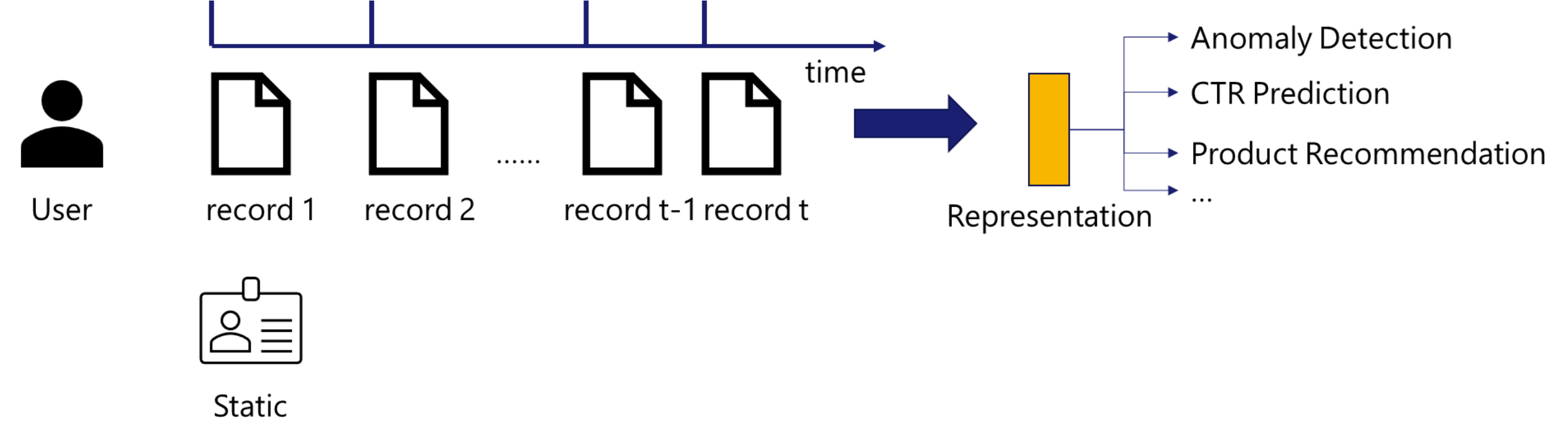}
\caption{\Tra{} sequence representation learning with both static and dynamic fields and time and order information. Given a sequence of \tras{}, the sequence is associated with both static fields and dynamic field. The order and time information of \tras{} are also given. The goal is to learn a useful representation of the sequence of \tras{} that can be used in downstream tasks.}
\label{fig:problem_def}
\end{figure}

In this study, we investigate the task of learning effective representations for sequential tabular datasets using a pre-training framework based on transformer models. The process is visually depicted in Figure \ref{fig:problem_def}, which uses a record sequence as an illustrative example. Each record represents a row in the tabular dataset, and records associated with the same identifier form a sequence. In Figure \ref{fig:problem_def}, all these records are associated with a specific user. The sequence comprises multiple records, each containing both static and dynamic fields. Static fields maintain consistent values throughout the sequence, while dynamic fields exhibit variations over time. Additionally, the order and time information of the records are provided.

More formally, given a sequential tabular dataset with $m$ \tras{} (rows), each \tra{} $x_i$ is composed of $n_{s}$ static fields and $n_{d}$ dynamic fields. An input, $X$, to our proposed method is represented as a windowed sequence of $l$ time-dependent rows (\tras{}) $x_i$,
\begin{align}
    X &= [x_{0}, x_{1}, ..., x_{l-1}] \\
    x_{i} &= \{ u_{i}^{s,0}, u_{i}^{s,1}, ..., u_{i}^{s,n_{s}-1}, u_{i}^{d,0}, u_{i}^{d,1}, ..., u_{i}^{d,n_{d}- 1} \}
\end{align}
where $l \ (\ll m)$ is the number of consecutive \tras{} selected with a window offset (or stride), $u_{i}^{s,j}$ are static fields, and $u_{i}^{d,k}$ are dynamic fields.
In the windowed sequence $X$, we assume that static fields always keep a constant value, that is $u_{i}^{s,j} = u^{s, j} \quad \textrm{for} \, i \in [0, l-1]$.

For each \tra{} $x_i$, we are provided with the creation time $t_i$. To simplify and maintain generality, we set the initial time $t_0$ as $0$. $t_i$ represents the time interval between the creation time of \tra{} $x_i$ and $x_0$. These \tra{} sequences can be effectively utilized for specific tasks, such as detecting anomalies in credit card transactions, by leveraging the time and order information of the transactions.

Prior to training a model for specific tasks, pre-training can be conducted to generate useful representations for \tra{} sequences. The pre-trained model can then be fine-tuned for specific downstream tasks. The primary objective of the pre-training task is to train a transformer-based model $f_\theta: X \rightarrow R$, where $R$ denotes a sequence of representations for \tra{}s,
\begin{align}
R = [r_{0}, r_{1}, ..., r_{l-1}]
\end{align}
where $r_{i} \in \mathbb{R}^{n_{e}}$. Here, $n_e$ is the dimension of learned representations.

\begin{figure*}
    \includegraphics[width=1\textwidth]{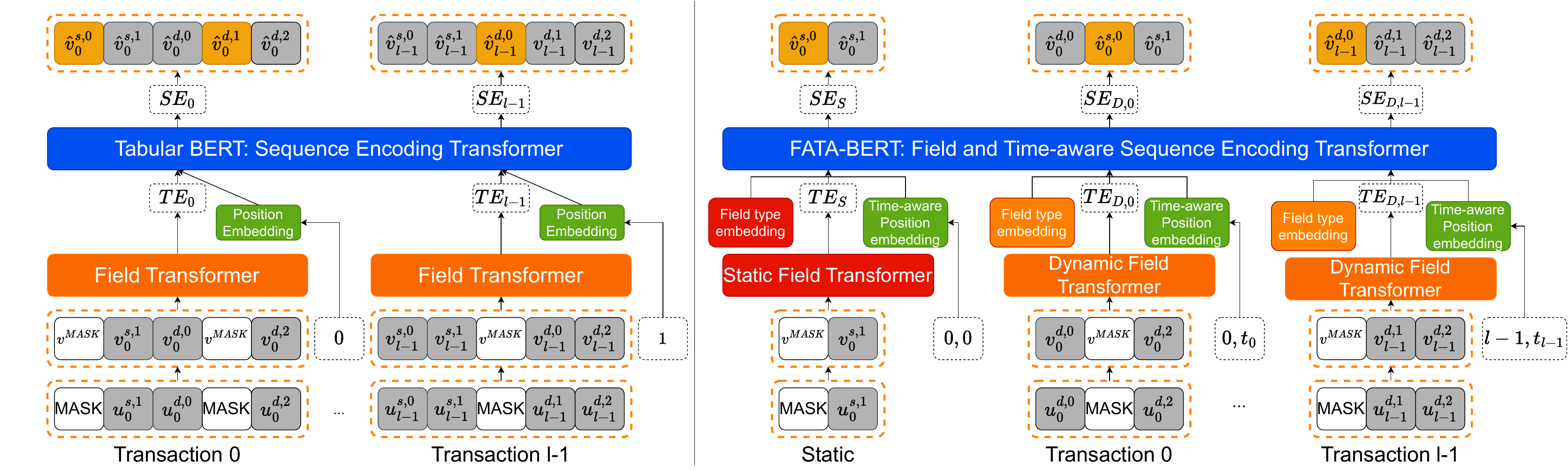}
    \caption{TabBERT framework (left) and \name{} framework (right). TabBERT framework processes \tra{} sequences in a hierarchical fashion, but it does not separate static and dynamic fields. Static fields are replicated over \tras{}. Also, position embedding in TabBERT does not consider time interval between \tras{}.  \name{} framework uses static and dynamic field transformers to process two types of fields separately. Static Fields are not replicated over \tras{}. Field type embedding distinguishes dynamic and static fields. Time-aware position embedding utilizes both time interval and sequence information.}
    \label{fig:trans}
\end{figure*}

\subsection{Previous Method: TabBERT}
Our proposed method, \name{}, is a variant of the TabBERT framework introduced in \cite{padhi2021tabular}. TabBERT was originally designed for modeling transaction sequences, where each \textit{transaction} can be considered as a \textit{\tra{}} in a general sequential tabular dataset. TabBERT defines each field on its own local vocabulary and quantizes numerical fields so that both categorical and numerical values can be represented by a finite vocabulary. 

TabBERT encodes the sequence of transactions in a hierarchical fashion. As shown in the left panel of Figure \ref{fig:trans}, TabBERT first uses a field transformer to process each transaction individually, creating transaction embeddings (\textit{TE}). Then these transaction embeddings are fed into the second-level transformer to create sequence embeddings (\textit{SE}). In this setting, the field transformer takes both static and dynamic field tokens as input. Because the value of each static field remains unchanged across transactions, static field tokens are replicated $l$ times in every input. This replication of static field tokens can lead to substantial computational resource usage if $n_{s}$ or $l$ is large. TabBERT uses the MLM procedure proposed in \cite{devlin2018bert} for pre-training. In MLM procedure, a certain percentage of the input tokens are masked at random, and then the model predicts those masked tokens. However, in TabBERT, if a static field token from one transaction is masked, the model can easily predict it by refernceing the same static field in other transactions. Consequently, the pre-training task becomes too simple, and the transformer fails to capture crucial relationships within transaction sequences.
Furthermore, TabBERT does not incorporate time interval information. While the position embedding layer injects transaction order information into the second-level transformer, the influence of time intervals is overlooked. This omission prevents TabBERT from fully capturing the temporal aspect of the data.

\subsection{Proposed Method: \name{}}
Our proposed method, \underline{F}ield \underline{A}nd \underline{T}ime-\underline{A}ware \underline{Trans}former for Sequential Tabular Data (\name{}),
is depicted in the right panel of Figure \ref{fig:trans}. \name{} consists of two levels: (1) At the first level, the static and dynamic fields are processed separately by the \emph{Static Field Transformer} and \emph{Dynamic Field Transformer} to generate \tra{} embeddings. (2) At the second level, a field type embedding distinguishes static and dynamic \tra{} embeddings, and a customized time-aware position embedding considers the impact of both time and order of \tras{} in a sequence. The \tra{} embedding, field type embedding, and time-aware position embedding are element-wise summed together and fed into the \emph{field and time-aware sequence encoding transformer (FATA-BERT)} to generate sequence embeddings. The generated sequence embeddings can then be used for downstream tasks.

\subsubsection{Dynamic Field Transformer And Static Field Transformer}
As shown in the right panel of Figure \ref{fig:trans}, \name{}  uses two field transformers to process static and dynamic fields separately. This distinction is crucial as static and dynamic fields represent different types of information in the record sequences. By processing them separately, \name{} reduces computational overhead and enables the model to capture important patterns within the sequences.
The transformation of raw field values into tokens follows a similar procedure to TabBERT. Both dynamic and static fields are tokenized. Subsequently, a random masking process is applied to some of these tokens, as described below:
\begin{subequations}
\label{eq:convert_mask}
\begin{align}
    v_{i}^{d, j} &= \text{ConvertToVocab}(u_{i}^{d, j}),\ \text{for}\ i \in [0, l-1], j \in [0, n_d - 1]\\
    \tilde{v}_{i}^{d, j}, \mathbb{I}_{i}^{d, j} &= \text{RandomMask}(v_{i}^{d, j}),\ \text{for}\ i\in [0, l-1], j \in [0, n_d - 1] \\
    v^{s, j} &= \text{ConvertToVocab}(u^{s, j}),\ \text{for}\ j \in [0, n_s - 1] \\
    \tilde{v}^{s, j}, \mathbb{I}^{s, j} &= \text{RandomMask}(v^{s, j}),\ \text{for}\ j \in [0, n_s - 1]
\end{align}    
\end{subequations}
In Equation \ref{eq:convert_mask}, $\mathbb{I}$ indicates whether a token is masked or not. If $\mathbb{I} = 0$, the token is replaced by [MASK]. This random masking procedure is for model pre-training only. 
After that, the Dynamic field transformer processes each \tra{} individually with only dynamic fields as input. The Static field transformer processes the static fields only once without replication. This helps to reduce computational overhead and prevents the model from seeing the masked static field tokens repeatedly. The Static field transformer ($f_{\theta_{sf}}$) produces the static \tra{} representation $TE_S$, and the Dynamic field transformer ($f_{\theta_{df}}$) produces dynamic \tra{} representations $TE_{D,0}, TE_{D,1}, ..., TE_{D,l-1}$ as follows:
\begin{subequations}
    \label{eq:trans_embeds}
    \begin{align}
        TE_S &= f_{\theta_{sf}}([\tilde{v}^{s, 0}, \tilde{v}^{s, 1}, ..., \tilde{v}^{s, n_s -1}])\\
        TE_{D,i} &= f_{\theta_{df}}([\tilde{v}_{i}^{d, 0}, \tilde{v}_{i}^{d, 1}, ..., \tilde{v}_{i}^{d, n_d -1}]),\ \text{for}\ i\in [0, l-1] 
    \end{align}
\end{subequations}
\subsubsection{FATA-BERT: Field and Time-Aware Sequence Encoding Transformer}
In the second level of the proposed architecture, we aim to capture the differences and relationships between static and dynamic fields, and utilize time interval information. To achieve this, we have customized the input representation for the second-level transformer, which we refer to as the Field and Time-Aware Sequence Encoding Transformer (FATA-BERT). FATA-BERT is mostly based on BERT-base \cite{devlin2018bert} architecture. However, differing from the original BERT and TabBERT models, the input representation of FATA-BERT is constructed by adding the corresponding \tra{}, field type, and time-aware position embeddings.

The field type embedding is a lookup table that stores embeddings of both static and dynamic field types. For $TE_S$, the corresponding field type embedding $FE_S$ is the static type embedding. On the other hand, for each $TE_{D,i}$, the corresponding field type embedding $FE_D$ is the dynamic type embedding. The embeddings of each field type are updated during the training procedure.

The time-aware position embedding $P(i)$ is defined as a vector of length $d$, where each element $p_{(i, j)}$ is defined by:
\begin{subequations}
\label{eq:time-aware}
\begin{align}
TPos(i) &= w_p * i + w_t * t_i + b \\
p_{(i, j)} &=
\begin{cases}
\sin{\frac{TPos(i)}{10000(\frac{2j}{d})}} & \text{if $j$ is even} \\
\cos{\frac{TPos(i)}{10000(\frac{2j}{d})}} & \text{if $j$ is odd}
\end{cases}
\end{align}
\end{subequations}
In this context, we use the variables $i$ and $j$ to represent the position index and time-aware position embedding dimension, respectively. The $TPos(i)$ function is used to merge information from both the position index $i$ and time interval $t_i$, which helps the model to capture time-aware position information through the trainable parameters $w_p$, $w_t$, and $b$, enabling the model to learn a flexible function.

It is important to note that for dynamic fields in the $i$th \tra{}, the corresponding time-aware position embedding is represented as $P(i)$. On the other hand, for static fields, we set $i$ to 0 and $t_i$ to $t_0$ to obtain the time-aware position embedding. This implies that the time-aware embedding for static fields is also represented as $P(0)$, which is the same as the time-aware embedding for dynamic fields in the 0th \tra{}.

The input representation for static fields ($IE_S$), and $i$th \tra{}'s dynamic fields ($IE_{D,i}$) are defined by the following formula:
\begin{subequations}
\label{eq:input_rep}
\begin{align}
    IE_S &= TE_S + P(0) + FE_S \\
    IE_{D, i} &= TE_{D, i} + P(i) + FE_D
\end{align}
\end{subequations}
We feed $[IE_S, IE_{D, 0}, IE_{D, 1}, ...,  IE_{D, l-1}]$ into FATA-BERT. The last layer of FATA-BERT generates a series of sequence embeddings $[SE_S, SE_{D,0}, SE_{D,1}, ... , SE_{D,l-1}]$ according to:
\begin{align}
    \begin{split}
    [SE_S, SE_{D,0}, SE_{D,1}, ..., SE_{D,l-1}] =  \\ f_{\theta_{fata}}([IE_S, IE_{D, 0}, IE_{D, 1}, ...,  IE_{D, l-1}])    
    \end{split}
\label{eq:fata-bert}
\end{align}
Here, $SE_S$ represents embeddings for the static fields, and $SE_{D, i}$ represents embeddings for the dynamic fields in the $i$th \tra{}. These sequence embeddings are then fed into the classification head ($f_{\theta_{cls}}$) to predict each token in the input \tra{} as follows: 
\begin{align}
\begin{split}
    [\Pr_{\theta_{cls}}(v^{s,0}), \Pr_{\theta_{cls}}(v^{s,1}), ..., \Pr_{\theta_{cls}}(v^{s,n_s -1}), \\ \Pr_{\theta_{cls}}(v_{0}^{d,0}), \Pr_{\theta_{cls}}(v_{0}^{d,1}), ... \Pr_{\theta_{cls}}(v_{0}^{d,n_d - 1}), \\ \Pr_{\theta_{cls}}(v_{1}^{d,0}), \Pr_{\theta_{cls}}(v_{1}^{d,1}), ..., \Pr_{\theta_{cls}}(v_{l - 1}^{d,n_d - 1}),] \\ = f_{\theta_{cls}}([SE_S, SE_{D,0}, SE_{D,1}, ..., SE_{D,l-1}])
\end{split}
\label{eq:token_predict}
\end{align}
In Equation \ref{eq:token_predict}, $\Pr_{\theta_{cls}}(v)$ represents the prediction probability for each token. Note that when we calculate the cross-entropy loss for the MLM task, as defined by the following formula, we only consider tokens that have been masked (masked indicator $\mathbb{I} = 0$ in our setting):

\label{eq:loss}
\begin{align}
    \ell = - \sum_{j = 0}^{n_s - 1}{(1 - \mathbb{I}^{s,j})\log{\Pr_{\theta_{cls}}(v^{s,j}) }} - \sum_{i = 0}^{l-1} \sum_{j = 0}^{n_d - 1}{(1 - \mathbb{I}_{i}^{d,j})\log{\Pr_{\theta_{cls}}(v_{i}^{d,j}) }}
\end{align}

\section{Experiments}

We compared the performance of \name{} with two powerful baseline models on three benchmark datasets. We also compared FATA-Trans to TabBERT which served as an inspiration for our study.

\subsection{Datasets and Tasks}
\paragraph{Synthetic Transaction Dataset - Transaction Anomaly Detection Task.}
This synthetic dataset was created by \cite{altman2021synthesizing,padhi2021tabular} for credit card transactions \footnote{\url{https://www.kaggle.com/datasets/ealtman2019/credit-card-transactions}}.  
It contains 24,386,900 transactions from 2,000 users’ 6,139 cards, covering a period of 1991 through 2020. Each transaction has attributes such as transaction time, merchant category code (MCC), transaction location, transaction type (chip, swipe or online), transaction amount, and an anomaly label indicating abnormal transactions. Some static fields were derived from these attributes such as average and standard deviation of transaction amounts, the most frequent MCC observed in the card transaction history, and the most frequent transaction type for each user.

The objective of this study was to predict whether the last transaction in a windowed sequence of card transactions is abnormal or not. To create these windowed sequences for each card, 10 consecutive transactions were combined in a time-dependent manner, as described in [30].

\paragraph{Amazon Product Reviews Dataset - Reviewer Rating Prediction Task.}
This dataset, collected by \cite{ni2019justifying}, consists of product reviews and metadata from Amazon \footnote{\url{https://nijianmo.github.io/amazon/index.html}}. Each review record includes fields such as review rating (range from 1 to 5), verification status, review time, reviewerID, asin (product ID). Static fields were created based on these attributes, including the average of historical ratings, the count of historical ratings, the percentage of low ratings (rating lower or equal to 3), and the percentage of high ratings (rating higher or equal to 4). This dataset contains 233.1 million reviews. In this study, we used two subsets under two categories: 5-core Movies and TV dataset (3,410,019 reviews) and 5-core Electronics dataset (6,739,590 reviews). Here, "5-core" means that all the remaining users and items in the dataset have at least 5 reviews each.

The objective of this task was to predict whether the last review in a reviewer's windowed sequence of reviews is a high rating or a low rating. The length of the windowed review sequence is set as 10.

\subsection{Compared Methods}
We compared \name{} with the following methods:

\paragraph{LightGBM}
LightGBM, developed by Microsoft\cite{ke2017lightgbm}, belongs to the family of gradient boosting trees \cite{friedman2001greedy} which have been the dominant solutions in various Kaggle and other industry competitions \cite{jannach2020deep} and used by researchers as both inspiration and the standard by which to compare model performance \cite{arik2021tabnet,fiedler2021simple,gorishniy2021revisiting, grinsztajn2022tree,shwartz2022tabular}. 
We chose LightGBM as our first baseline because of its superior performance in modeling tabular data.

\paragraph{RNN}
Recurrent neural networks (RNNs) have demonstrated remarkable performance in domains involving sequential data and the availability of user identifiers (e.g., credit card numbers or login IDs) \cite{lipton2015critical,zhang2021transaction,branco2020interleaved,li2018transaction,hidasi2015session,zhang2014sequential}. Thus, we selected RNNs as our second baseline due to the sequential nature of our data and the presence of user identifiers.
Specifically, we chose the gated recurrent unit (GRU) RNN \cite{cho2014learning}, which uses less memory and is faster to train compared with another commonly used RNN architecture known as long short-term memory (LSTM) \cite{hochreiter1997long}. 

\paragraph{TabBERT}
The TabBERT model, proposed by \cite{padhi2021tabular}, follows a hierarchical structure, where the first-level field transformer generates transaction embeddings from individual transactions. These embeddings are then used as input for the second-level transformer to generate sequence embeddings. However, it is important to note that TabBERT does not take into account time interval information when computing position embeddings, 
and it does not distinguish between static and dynamic fields. To ensure comparability, we included the time interval as a dynamic field in TabBERT, allowing it to leverage transaction time information.

\subsection{Experimental Setup}
\paragraph{Data Preprocessing}
For the synthetic credit card transaction dataset, we divided all transactions occurring before 2018 into separate training and validation datasets. Transactions that occurred after 2018 were designated as the test (holdout) dataset. Within the training and validation datasets, we randomly selected 84\% of the transaction sequences for training purposes, while the remaining sequences were used for validation.

For the Amazon product review dataset, we followed a similar approach as described in  \cite{chen2022denoising, kang2018self, li2020time}. For each user, we included the last 10 reviews in the test dataset. The reviews from the 11th to the second-to-last review were placed in the validation dataset, and all the reviews before the second-to-last review were assigned to the training dataset. In cases where the sequence length was less than 10, we added a special token to the left of the sequence repeatedly until the length reached 10.

We created transaction/review sequences as sliding windows of 10 \tra{}s, with a stride of 5 in training data and validation data, and a stride of 1 in test data. We followed the same procedure outlined in \cite{padhi2021tabular} to quantize numerical features and generate vocabularies.

\paragraph{Model Pre-training and Downstream Task Training.}

For our experiments, we utilized TabBERT and FATA-Trans as our pre-trained models. When working with the synthetic transaction dataset, 
we  intentionally excluded the label column, which indicates whether a transaction is abnormal, to avoid any leakage of target information during the pre-training phase \cite{kaufman2012leakage}. However, for the Amazon product review dataset, we included the label column representing a reviewer's rating on a product. This inclusion was based on the assumption that previous ratings can be informative for predicting future ratings. During our experiments, we followed a similar procedure as described in \cite{padhi2021tabular,devlin2018bert}. We randomly masked 15\% of the static field tokens and 15\% of the dynamic field tokens in each sequence. These masked tokens were replaced with the [MASK] token 80\% of the time, with random tokens 10\% of the time, and left unchanged 10\% of the time. The model learned to restore these masked tokens using the cross-entropy loss. We pre-trained all the models for three epochs, using the same parameter settings in \cite{padhi2021tabular}.

After pre-training, we applied a linear layer as the classification head, taking the concatenated sequence embeddings as input. For the synthetic transaction dataset, we down-sampled the normal transaction sequences in the training dataset for the classification task training procedure. This was done to make the ratio of normal to abnormal transactions 20:1 since abnormal transactions are extremely rare. Note that during the pre-training procedures, models used the whole training dataset. For the Amazon product review datasets, we masked the rating column of the last review in each sequence to prevent target leaking \cite{kaufman2012leakage}. This ensured that the model did not have access to the prediction label during training. We trained both models for 20 epochs and employed early-stop criteria when the AUC (area under the receiver operating characteristic curve) score for the validation dataset did not improve over three consecutive evaluation steps.

It should be noted that both LightGBM and RNN models were
directly trained for the classification tasks without pre-training. Again, for the synthetic transaction dataset, we down-sampled the normal transaction sequences in the training dataset to achieve a 20:1 ratio between normal and abnormal transactions. For the LightGBM model, we utilized the LightGBM Python library\footnote{\url{https://pypi.org/project/lightgbm/}}. We primarily used the default parameter settings recommended by the package but made adjustments to the learning rate and number of boosting rounds based on the AUC score of the validation dataset.

The RNN model was trained using the Adam optimizer \cite{kingma2014adam} with a fixed learning rate of 0.001. A batch size of 64 was used for the synthetic transaction dataset, while a batch size of 128 was used for the two Amazon product review datasets. The model was trained for 100 epochs, and early-stop decisions were determined by the AUC score of the validation dataset. During our experimentation, we evaluated both one-layer and two-layer GRU networks and found that they exhibited nearly identical performance. Therefore, we present the results obtained from a one-layer GRU network with 256 hidden nodes.

\begin{table}[ht]\small
\caption{Performance comparison on three datasets: Synthetic Transaction Dataset, Amazon Movies and TV, and Amazon Electronics.}
\begin{tabular}{p{0.60\linewidth}  p{0.38\linewidth}}
    \toprule
    Methods & AUC \\
    \midrule
    \multicolumn{2}{l}{\textbf{Synthetic Transaction}} \\
    \midrule
    LightGBM                                       & 0.9624                \\
    RNN                                            & 0.9866                \\
    TabBERT                                        & 0.9985                \\
    \name{}                                     & \textbf{0.9992}                \\
    \midrule
    \multicolumn{2}{l}{\textbf{Amazon Movies and TV}} \\
    \midrule
    LightGBM                                       & 0.7593                \\
    RNN                                            & 0.7634                \\
    TabBERT                                        & 0.7964                \\
    \name{}                                     & \textbf{0.8057}                \\
    \midrule
    \multicolumn{2}{l}{\textbf{Amazon Electronics}} \\
    \midrule
    LightGBM                                       & 0.6962                \\
    RNN                                            & 0.7040                \\
    TabBERT                                        & 0.7098                \\
    \name{}                                     & \textbf{0.7206}                \\
    \bottomrule
    
\end{tabular}
    \label{tab:all-task}
\end{table}

\subsection{Experimental Results}

We evaluated AUC scores on the testing dataset for both the transaction anomaly detection and review rating prediction tasks. As depicted in Table \ref{tab:all-task}, our method consistently outperforms other methods across all three datasets. This indicates that \name{} effectively captures more precise user behavior patterns by leveraging the time interval and field-type information incorporated within the specially designed embedding and transformer layers.

\begin{table}[ht]\small
\caption{Pre-training time comparison on three datasets: Synthetic Transaction, Amazon Movies and TV, and Amazon Electronics. We used batch size 64 for the Synthetic Transaction dataset and batch size 128 for the other two datasets.}
\begin{tabular}{p{0.60\linewidth}  p{0.38\linewidth}}
\toprule
Methods                                    & Pretraining Time                                 \\ 
\midrule
\multicolumn{2}{l}{\textbf{Synthetic Transaction}}                                \\ 
\midrule
TabBERT                                    & 3d11h                               \\
\name{}                                 & \textbf{2d12h}                       \\ 
\midrule
\multicolumn{2}{l}{\textbf{Amazon Movies and TV}} \\ 
\midrule
TabBERT                                    & 11h                                  \\
\name{}                                 & \textbf{6h33m}                       \\ 
\midrule
\multicolumn{2}{l}{\textbf{Amazon Electronics}}   \\ 
\midrule
TabBERT                                    & 2d13h                                \\
\name{}                                 & \textbf{1d3h}                        \\
\bottomrule
\end{tabular}
    \label{tab:pretraining-time}
\end{table}

We conducted a comparison of the pre-training time between TabBERT and \name{}. Both models were implemented using PyTorch \cite{paszke2019pytorch} and pre-trained on a single NVIDIA Tesla A100 GPU. The batch size was set to 64 for the synthetic transaction dataset and 128 for the other two datasets. According to Table \ref{tab:pretraining-time}, our method demonstrates significantly shorter pre-training times compared to TabBERT. This is attributed to the fact that \name{} avoids redundant repetition of static fields in the sequence and only inputs them into the static-field transformer. As a result, this approach reduces memory usage and substantially saves training time.

\begin{table}[ht]\small
\caption{Performance of the proposed \name{} and its variations on three datasets: Synthetic Transaction, Amazon Movies and TV, and Amazon Electronics.}
\begin{tabular}{p{0.70\linewidth}  p{0.28\linewidth}}
    \toprule
    Methods & AUC \\
    \midrule
    \multicolumn{2}{l}{\textbf{Synthetic Transaction}} \\
    \midrule
    \name{} (w/o time-aware position embedding)   & 0.9982                \\
    \name{} (w/o field type aware design)         & 0.9966                \\
    \name{} (w/o pretraining)                     & 0.9970                 \\
    \name{}                                       & \textbf{0.9992}                \\
    \midrule
    \multicolumn{2}{l}{\textbf{Amazon Movies and TV}} \\
    \midrule
    \name{} (w/o time-aware position embedding)   & 0.8036                \\
    \name{} (w/o field type aware design)         & 0.8038                \\
    \name{} (w/o pretraining)                    & 0.7819                \\
    \name{}                                     & \textbf{0.8057}                \\
    \midrule
    \multicolumn{2}{l}{\textbf{Amazon Electronics}} \\
    \midrule
    \name{} (w/o time-aware position embedding)   & 0.7108                \\
    \name{} (w/o field type aware design)         & 0.7174                \\
    \name{} (w/o pretraining)                    & 0.7072                \\
    \name{}                                     & \textbf{0.7206}                \\
    \bottomrule
    
\end{tabular}
    \label{tab:ablation}
\end{table}

\subsection{Ablation Study} 
We designed two variations of our proposed method to assess the  impact of our customized time-aware position embedding, static field transformer, and field-type embedding. Each variant differs from \name{} in either position embedding or field transformer and field-type embedding. We also compared \name{} with a variant that did not utilize pre-training, but instead directly trained on the downstream classification task.

\paragraph{\name{} (w/o time-aware position embedding)} 
This variation replaced the time-aware position embedding with the regular position embedding proposed in \cite{devlin2018bert}. It still used the time interval as a dynamic field to capture the time information.

\paragraph{\name{} (w/o field type aware design)} 
This variation removed the static field transformer and field type embedding. It replicated the static fields in every \tra{} in a sequence as TabBERT did.

\paragraph{\name{} (w/o pre-training)} 
This variation skipped the pre-training step and instead directly trained on the downstream task.

Table \ref{tab:ablation} shows the experimental results on all three datasets. \name{} achieves better results against all three variations, which demonstrates that the time-aware position embeddings, static field transformer, and field-type embedding can better exploit the time interval information and better learn the latent patterns across fields and \tras{}. Moreover, the pre-training procedure can help the method obtain more useful representations for the downstream tasks.

\begin{figure*}
    \medskip
    \begin{subfigure}[t]{0.3\textwidth}
        \centering
         \includegraphics[width=\textwidth]{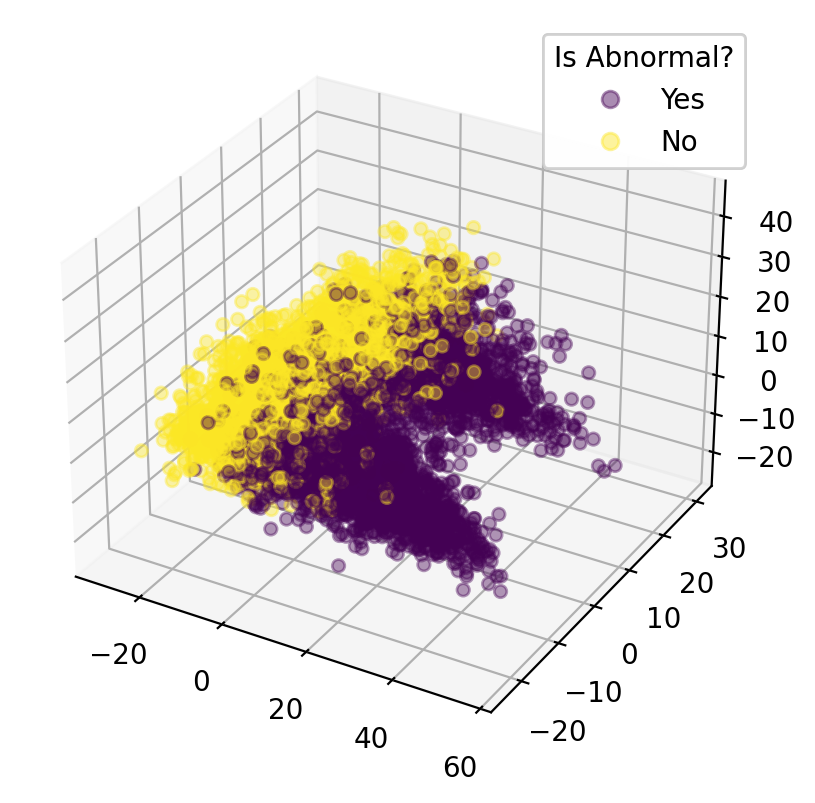}
         \caption{Scatter plot for abnormal transactions in the Synthetic Transaction Dataset. Each dot represents the first three principal components of the concatenated sequence embeddings of a windowed transaction sequence. The color of the dot indicates whether the last transaction in the sequence is abnormal or not.}
         \label{fig:whole_seq_abnormal}
    \end{subfigure}\hfill
    \begin{subfigure}[t]{0.3\textwidth}
        \centering
        \includegraphics[width=\textwidth]{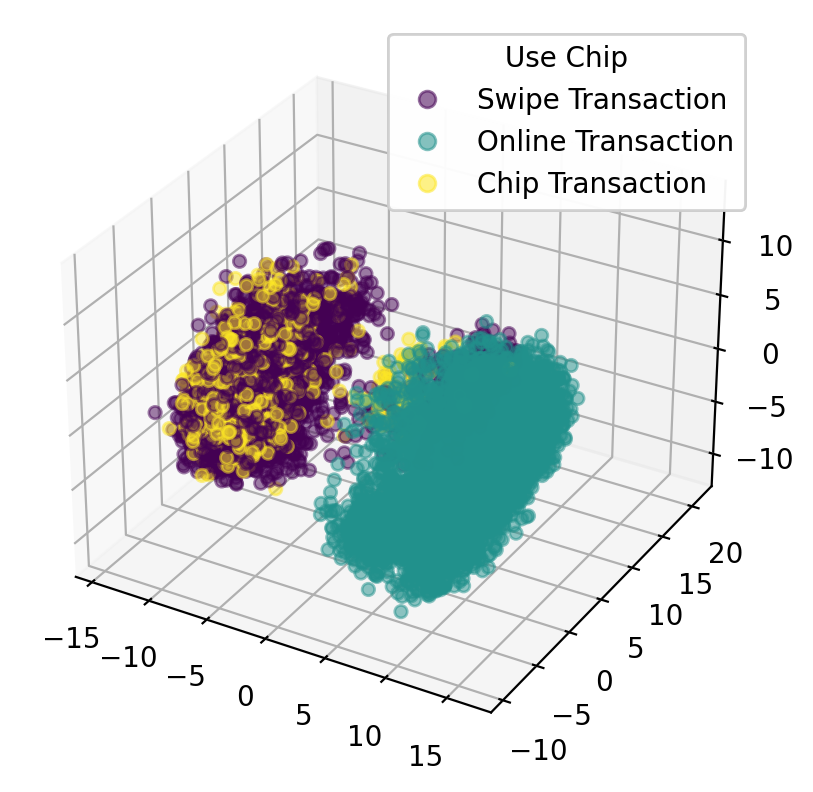}
        \caption{Scatter plot for the transaction types in the Synthetic Transaction Dataset. Each dot represents the first three principal components of a transaction's sequence embedding. The color of the dot represents transaction types: swipe, online, or chip.}
        \label{fig:last_seq_chip}
    \end{subfigure}\hfill
    \begin{subfigure}[t]{0.3\textwidth}
        \centering
        \includegraphics[width=\textwidth]{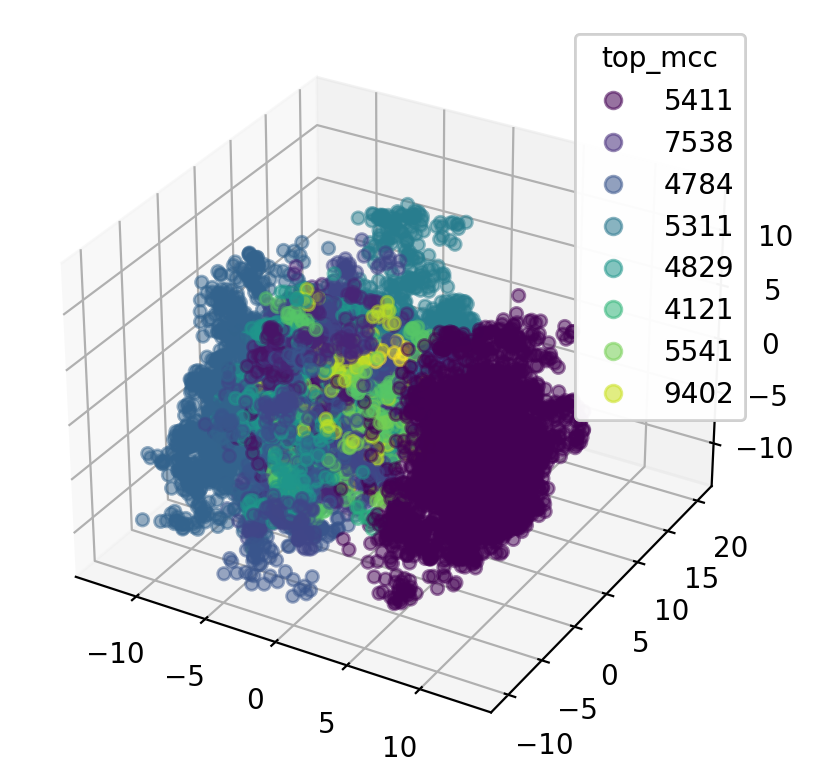}
        \caption{Scatter plot for top MCCs in the Synthetic Transaction Dataset. Each dot represents the first three principal components of a static fields' sequence embedding. Embeddings are generated by pre-trained FATA-Trans. The color of the dot represents the most frequent MCC in a card's transaction history.}
        \label{fig:static_mcc}
    \end{subfigure}\hfill
    \begin{subfigure}[t]{0.3\textwidth}
        \centering
        \includegraphics[width=\textwidth]{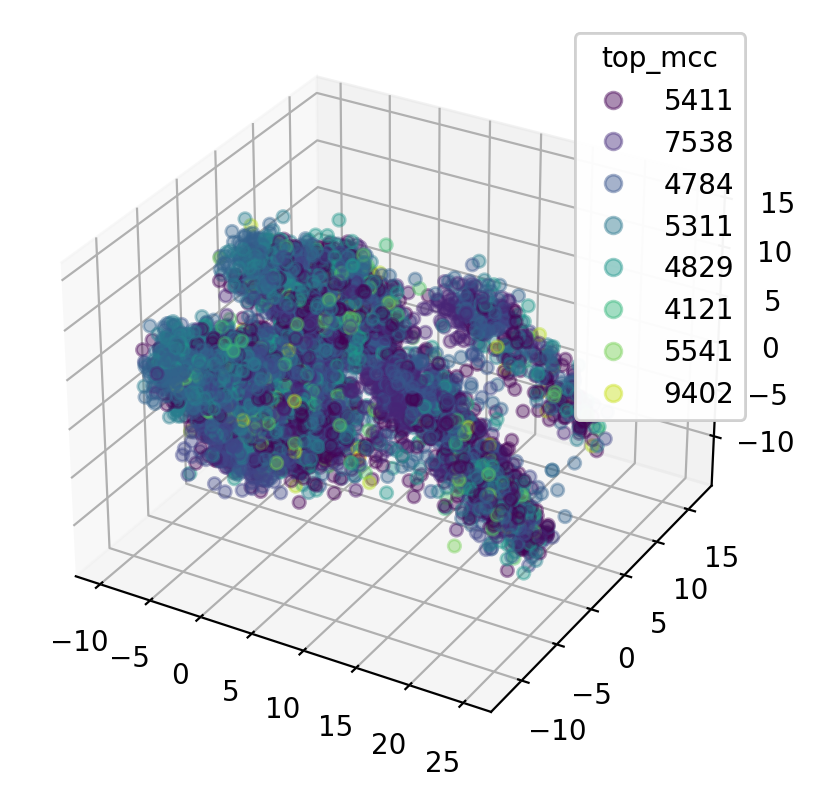}
        \caption{Scatter plot for top MCCs in the Synthetic Transaction Dataset. Each dot represents the first three principal components of a transaction's sequence embedding. Embeddings are generated by pre-trained TabBERT. The color of the dot represents the most frequent MCC in a card's transaction history.}
        \label{fig:normal_mcc_tabbert}
    \end{subfigure}\hfill
    \begin{subfigure}[t]{0.3\textwidth}
        \centering
        \includegraphics[width=\linewidth]{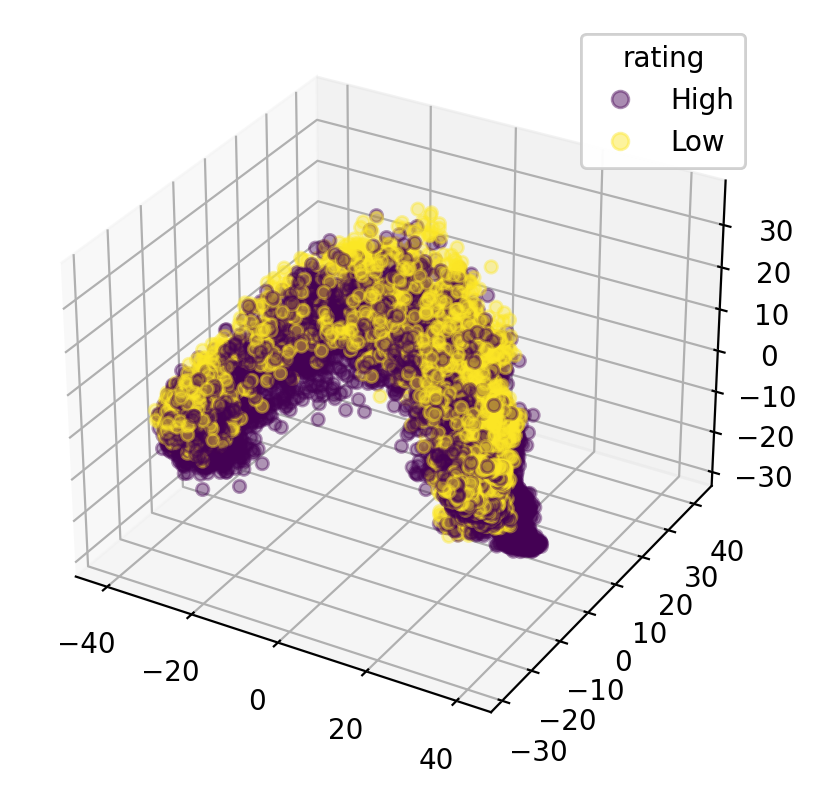}
        \caption{Scatter plot for reviewer ratings in Amazon Electronics. Each dot represents the first three principal components of the concatenated sequence embeddings of a windowed review sequence. Embeddings are generated by pre-trained \name{}. The color of the dot represents if the last rating in the sequence is high (>=4) or low (<=3).}
        \label{fig:whole_seq_electronic_overall}
    \end{subfigure}\hfill
    \begin{subfigure}[t]{0.3\textwidth}
        \centering
        \includegraphics[width=\linewidth]{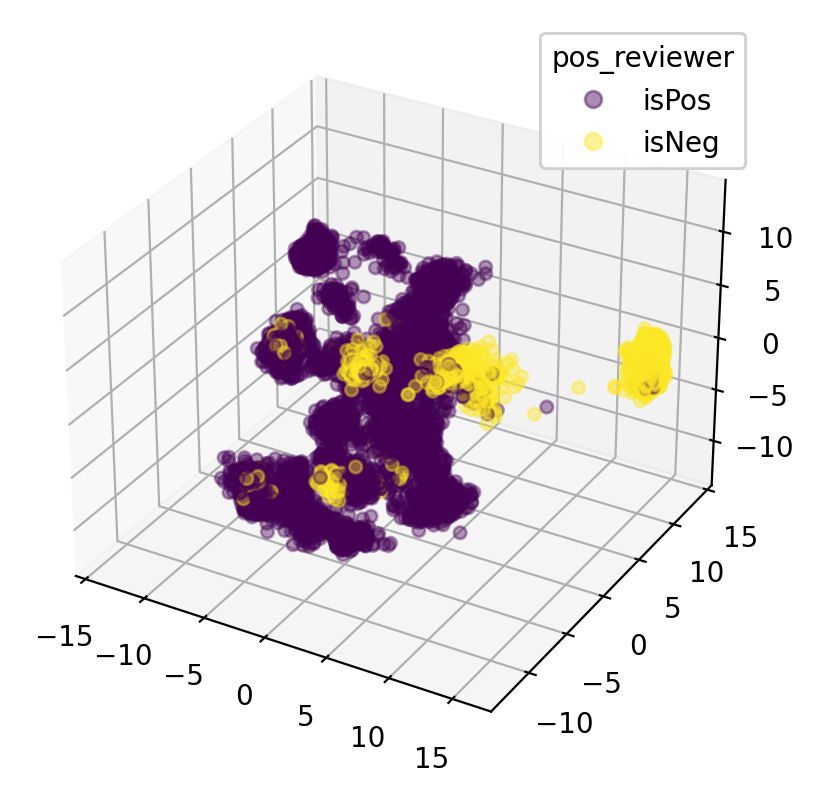}
        \caption{Scatter plot for high or low rating reviewer in Amazon Movies and TV. Each dot represents the first three principal components of a static fields' sequence embedding. Embeddings are generated by pre-trained \name{}. The color of the dot represents if a reviewer gave more high rating (>=4) or low rating (<=3) in the reviewer's review history.}
        \label{fig:static_pos_reviewer}
    \end{subfigure}
\end{figure*}

\subsection{Representation Visualization}

We used the learned sequence embeddings from FATA-Trans, which were not fine-tuned for any particular downstream task, to explore the insights captured by our model. To analyze these embeddings, we applied Principal Component Analysis (PCA) and generated 3D plots using the first three principal components. To create the plot, we either concatenated the sequence embeddings within each window or used a single sequence embedding and then applied PCA.

In Figure \ref{fig:whole_seq_abnormal}, we present the distribution of anomaly labels for the synthetic transaction dataset. Each point in the figure represents a windowed transaction sequence. Notably, we observe a clear separation between abnormal and normal sequences, suggesting that our learned representation has captured this information even without explicitly training an anomaly detection model. Figure \ref{fig:last_seq_chip} presents the distribution of transaction types within the 3D space. Each point represents a transaction \tra{}. Again, we can observe a near-perfect separation between the two common transaction types: online and offline (or point-of-sale) involving swipe and chip transactions.
Figure \ref{fig:static_mcc} illustrates the distribution of several top MCCs (the most frequent MCCs in a card's transaction history). Each point represents the sequence embedding of the static fields within a windowed transaction sequence. Notably, we can observe significant separation among these MCCs. It's important to note that transaction type is a dynamic field, while the top MCC is a static field. Additionally, we performed the same visualization task using the sequence embeddings learned by the pre-trained TabBERT model. Figure \ref{fig:normal_mcc_tabbert} displays the distribution of transaction sequence embeddings learned by the TabBERT model. It is worth noting that TabBERT does not differentiate between static and dynamic fields. When comparing Figure \ref{fig:static_mcc} (FATA-Trans) and Figure \ref{fig:normal_mcc_tabbert} (TabBERT), we observe that the embeddings from TabBERT do not effectively separate the top MCCs, which is a static field. This indicates that FATA-Trans excels at capturing information from static fields compared to the TabBERT model.

In Figure \ref{fig:whole_seq_electronic_overall}, we visualize the distribution of reviewer ratings for the Amazon Electronics dataset. Each point represents a windowed review sequence, and we can observe a clear separation between high and low rating sequences. This indicates that our method successfully captures the underlying patterns associated with different rating levels.
In Figure \ref{fig:static_pos_reviewer}, we illustrate the separation between positive and negative reviewers based on their review history in the Amazon Movie and TV dataset. The feature "is-pos" is a derived feature that is created based on a user's review history. It serves as a static field and reflects a user's overall preference. Each point in the figure corresponds to the sequence embedding of the static fields within a windowed review sequence. Notably, we can observe a significant separation between positive and negative reviewers, indicating that our approach effectively extracts patterns from both static and dynamic fields within record sequences.

These figures collectively demonstrate the capability of our method to successfully capture and extract meaningful information from both static and dynamic fields within sequences.

\section{Related Work}
Sequential tabular data is ubiquitous across many industries. Different from attributed sequences~\cite{zhuang2019attributed}, where each attribute is static, sequential tabular data has both static and dynamic feature fields. Recently, there has been growing interest in applying transformer-based models \cite{devlin2018bert,vaswani2017attention} to tabular data. For example, SAINT\cite{somepalli2021saint} introduces an inter-sample attention strategy for modeling tabular datasets.  TabNet\cite{arik2021tabnet} uses a sequential attention mechanism to choose a subset of semantically meaningful features to process at each decision step. TabTransformer\cite{huang2020tabtransformer} uses a transformer encoder to learn contextual embeddings on categorical features. AutoInt\cite{song2019autoint} automatically learns high-order feature interactions for CTR prediction using a self-attentive neural network. Shwartz-Ziv and Armon\cite{shwartz2022tabular}, Gorishniy et al. \cite{gorishniy2021revisiting,gorishniy2022embeddings}, Rubachev et al. \cite{rubachev2022revisiting}, and Levin et al. \cite{levin2022transfer} conduct in-depth studies comparing the main families of deep learning architectures against gradient boosting trees. Borisov et al \cite{borisov2022deep} and Badaro et al. \cite{badaro2022transformers} present extensive surveys in this field. However, most of these works are focused on non-sequential tabular data where rows in a table are independent and there are no temporal dependencies between the rows.

There is another line of research in sequential recommendation which leverages the sequential nature of a user’s behavior to make better recommendations. Early work utilized RNN models and achieved state-of-art performance\cite{hidasi2015session,yu2016dynamic,zhang2014sequential}. Recently, transformer-based models have become a proliferated approach. For example, SASRec\cite{kang2018self} uses a self-attention mechanism combined with position embeddings to learn relevant items based on a user’s past purchased items. TiSASRec\cite{li2020time} incorporates relative time intervals between any two items in a sequence into a self-attention mechanism to predict the next item that a user is likely to engage with. BERT4Rec\cite{sun2019bert4rec} applies bidirectional attention to capture a users’ sequential behavioral patterns. TLSRec\cite{chen2022time} simultaneously models the global stability and local fluctuation of a user’s preference with a hierarchical attention network. Rec-Denosier\cite{chen2022denoising} adaptively eliminates the noisy items during the training process to remove irrelevant information in a user’s behavior sequence. Transformers4Rec\cite{de2021transformers4rec} performs an empirical analysis with broad experiments of various transformer architectures for the task of sequential recommendation. Despite encouraging performance,
a common limitation of these approaches is their exclusive focus on item IDs within a univariate sequence, disregarding other valuable information associated with items and users, such as item category, item popularity, user past comments and ratings, and more.

Several studies have addressed the limitation of exclusively relying on item IDs by incorporating other valuable information associated with items and users. For instance, FDSA \cite{zhang2019feature} introduces a feature-level self-attention block to integrate detailed attribute information about items. BST \cite{chen2019behavior} combines both item IDs and category IDs to construct user behavior sequences, which are then fed into a transformer layer. SSE-PT\cite{wu2020sse} incorporates user embeddings into a self-attentive neural network to personalize the transformer model. $S^3$-Rec\cite{zhou2020s3} utilizes mutual information maximization within a self-attentive architecture to capture correlations among attributes, items, subsequences, and sequences. TabBERT\cite{padhi2021tabular}, from which we got inspiration, provides a comprehensive framework for modeling multivariate sequential tabular data. However, these approaches overlook the distinction between static and dynamic fields and do not account for time interval information in position embedding.

\section{Conclusion}
In this paper, we present \name{}, a novel Field- and Time-Aware Transformer for modeling \tra{} sequences in sequential tabular data. Compared to previous works, \name{} has a special design to process static and dynamic fields separately, and the time interval information is also incorporated into time-aware position embedding. We show that the representations learned by \name{} provide consistent performance gain in both transaction anomaly detection and product review rating prediction tasks, achieved with substantially less training time. Visualization figures also show that \name{} can capture important information from both static and dynamic fields.

\bibliographystyle{ACM-Reference-Format}
\balance
\bibliography{paper_bib}


\end{document}